\documentclass{article}

\usepackage[preprint, nonatbib]{neurips_2023}

\usepackage[utf8]{inputenc} 
\usepackage[T1]{fontenc}    
\usepackage{hyperref}       
\usepackage{url}            
\usepackage{booktabs}       
\usepackage{amsfonts}       
\usepackage{nicefrac}       
\usepackage{microtype}      
\usepackage{xcolor}         
\usepackage{amsmath}
\usepackage{wrapfig}
\usepackage{graphicx}
\usepackage{enumitem}
\usepackage{longtable}
\usepackage[skip=1pt]{caption}

\hypersetup{
	colorlinks,
	citecolor={red!40!gray},
	linkcolor={blue!40!gray},
	urlcolor={blue!70!gray}
}

\definecolor{darkgreen}{RGB}{50, 168, 82}

\title{Preference Optimization for Molecular Language Models}

\author{%
  Ryan Park\\
  Harmonic Discovery Inc.\\
  Stanford University\\
  \texttt{rypark@stanford.edu}
  \And
  Ryan Theisen\\
  Harmonic Discovery Inc.\\
  \texttt{ryan@harmonicdiscovery.com}
  \And 
  Navriti Sahni\\
  Harmonic Discovery Inc.\\
  \texttt{navriti@harmonicdiscovery.com}
  \And 
  Marcel Patek\\
  Harmonic Discovery Inc.\\
  \texttt{marcel@harmonicdiscovery.com}
  \And 
  Anna Cicho\'{n}ska\\
  Harmonic Discovery Inc.\\
  \texttt{anna@harmonicdiscovery.com}
  \And 
  Rayees Rahman\\
  Harmonic Discovery Inc.\\
  \texttt{rayees@harmonicdiscovery.com}
}

\begin{document}

\maketitle

\begin{abstract}
  Molecular language modeling is an effective approach to generating novel chemical structures. However, these models do not \emph{a priori} encode certain preferences a chemist may desire. We investigate the use of fine-tuning using Direct Preference Optimization to better align generated molecules with chemist preferences. Our findings suggest that this approach is simple, efficient, and highly effective.   
\end{abstract}

\section{Introduction}
\label{sec:intro}
In recent years, molecular language models have proven remarkably effective for molecule generation tasks. Such approaches utilize string representations of molecules, such as SMILES \cite{SMILES} or SELFIES \cite{SELFIES}, together with standard language modeling architectures to learn and sample from a distribution over chemical structures. These models can then be leveraged for tasks such as drug and material design. For example, the MOSES benchmarking suite finds that a simple LSTM architecture trained on drug-like molecules, represented as SMILES, achieves at or near state-of-the-art performance across a variety of metrics measuring drug-likeness, synthesizability, diversity, and novelty \cite{MOSES}.

However, molecular language models do not by default encode all properties required for use in many practical settings. For example, a medicinal chemist may be interested in chemical structures containing only a particular substructure, without certain reactive groups, or having other properties such as binding affinity against a given target of interest. While it is possible to perform post-hoc filtering or optimization for these properties after sampling (e.g., using Monte Carlo Tree Search \cite{ChemTS}), it is desirable to have models that can \emph{a priori} encode arbitrary preferences as desired by a user. 

In this work, we explore the use of Direct Preference Optimization (DPO) \cite{DPO} to encode preferences directly into molecular language models via fine-tuning. Since many of the properties of interest can be directly and efficiently computed (or estimated), it is possible to cheaply generate large labeled, synthetic datasets for this task using a pre-trained language model. We show that this approach, together with the DPO fine-tuning strategy, enables significant improvement in the quality of generated molecules. 
\section{Setup and background}

\subsection{Preference optimization with DPO}
\label{sec:dpo}
\begin{figure}
    \centering
    \includegraphics[scale=0.4]{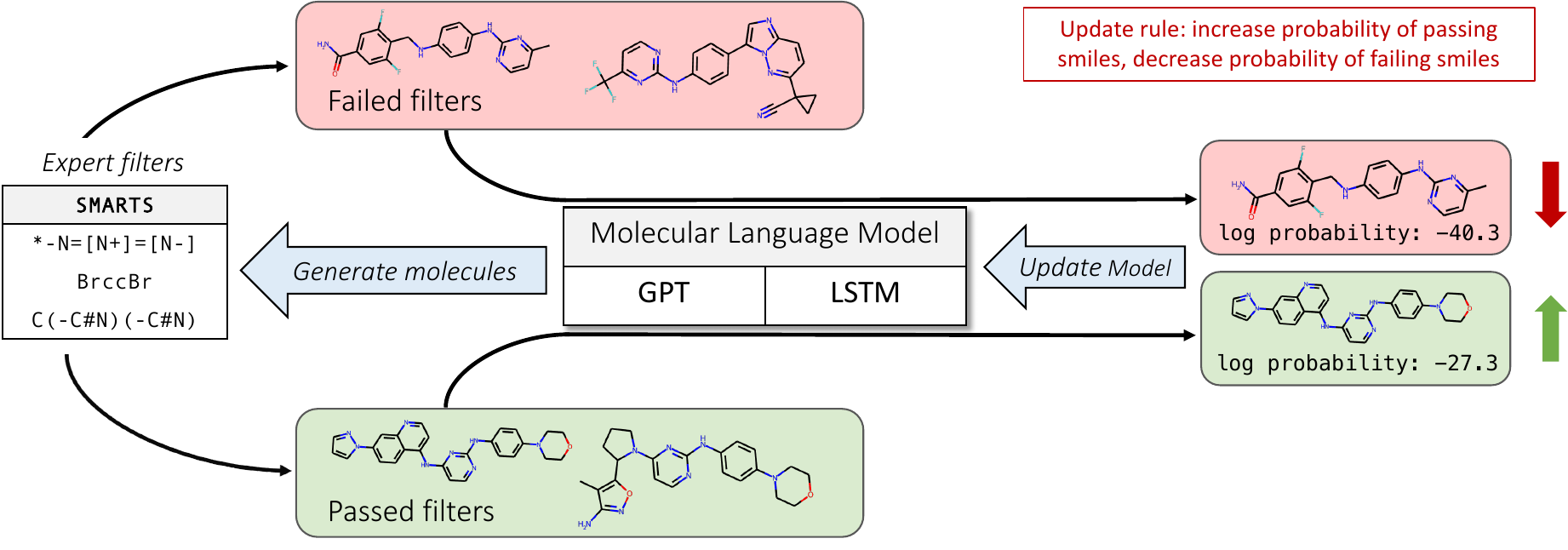}
    \vspace{5mm}
    \caption{Schematic representation of using DPO to fine-tune a molecular language model to produce molecules that pass medicinal chemist filters.}
    \label{fig:schematic}
\end{figure}

DPO was recently introduced as an effective alternative to reinforcement learning (RL) for fine-tuning language models using ordered preference data \cite{DPO}. Importantly, unlike RL, DPO does not require learning a separate reward model prior to fine-tuning, and instead facilitates downstream training of a language model directly from preference data. Given a fixed, pre-trained reference model $\pi_{\text{ref}}(s)$, which assigns a probability to a sequence $s$, and ranked pairs $s_p \succ s_n$, the DPO objective can be written as 

\begin{align*}
\mathcal{L}_{\text{DPO}}(\pi_{\theta}; \pi_{\text{ref}}) =  -\mathbb{E}_{(s_p,s_n)}\left[\log \sigma \left(\beta \log \frac{\pi_{\theta}(s_p)}{\pi_{\text{ref}}(s_p)} - \beta \log \frac{\pi_{\theta}(s_n)}{\pi_{\text{ref}}(s_n)} \right)\right],
\end{align*}

where $\pi_{\theta}$ is the model to be fine-tuned, and $\beta$ is a hyper-parameter controlling the extent to which the fine-tuned model deviates from the reference model. Intuitively, the DPO objective shifts probability mass away from negative sequences $s_n$ towards positive sequences $s_p$, while not deviating too much from the reference model.

The procedure for fine-tuning using DPO is as follows: generate sequences $s_1,\dots,s_N$ from the reference model $\pi_{\text{ref}}$, and rank them pairwise. One practical way to do this is to assign a numerical score $y_1,\dots,y_N$ to each sequence, and construct a positive/negative pair $(s_i,s_j)$ such that $y_i > y_j$. This is the approach we take for training with DPO in the present work. 

\subsection{Experimental setup}
\label{sec:setup}

Here, we briefly describe the experimental setup used in the present study. We include additional details in Appendix \ref{sec:additional-experimental-details}; code to reproduce our results is available at \href{https://github.com/Harmonic-Discovery/pref-opt-for-mols}{\texttt{https://github.com/Harmonic-Discovery/pref-opt-for-mols}}.

\paragraph{Model architectures.}
We perform experiments on two language modeling architectures: a generative pre-trained transformer architecture (GPT) \cite{GPT}, and a simple but popular LSTM-based architecture used in \cite{CharRNN}, closely based on the implementation in the MOSES benchmark suite \cite{MOSES}. We refer to the GPT model pre-trained on MOSES as \texttt{smiles-gpt-base} and the pre-trained LSTM model as \texttt{smiles-rnn-base}.

\paragraph{Data.} For pre-training, we use the MOSES benchmark dataset \cite{MOSES}, which consists of 1.9M unique SMILES strings extracted from the ZINC Clean Leads database. For fine-tuning, we also query molecules from the ChEMBL database \cite{ChEMBL}.

\paragraph{Metrics.} To evaluate models, we use several standard metrics. Specifically:
\begin{itemize}
    \item $\textbf{FracValid}$: the fraction of generated molecules that represent valid chemical structures (i.e., can be parsed from SMILES strings by the Python package \texttt{RDKit}).
    \item $\textbf{FracUnique}$: the fraction of valid generated molecules that are unique. 
    \item $\textbf{IntDiv}$: the internal diversity of a set of molecules, defined by $$1-\binom{M}{2}^{-1}\sum_{i,j;\,i\neq j} T_c(m_i,m_j),$$ where $T_c$ is the Tanimoto similarity, computed using ECFP4 fingerprints of molecules $m_i$, and $M$ is the number of molecules.
\end{itemize}
Throughout, all metrics we report are calculated based on 10,000 SMILES strings sampled from the relevant model.

\paragraph{Training.} All training was performed using a single NVIDIA A5000 GPU with 24GB of memory.

\section{Experiments}
\label{sec:experiments}
We present results from two different experiments using DPO to fine-tune molecular language models. Our focus here is on tasks specifically relevant to drug discovery and, moreover, we exclusively rely on feedback that can be computed automatically.

\subsection{Molecule filtering}
\label{sec:MCF-opt}
\begin{table}[]
\begin{center}
\resizebox{\textwidth}{!}{\begin{tabular}{|l|l|l|l|l|}
\hline
\textit{Model}                       & \textbf{FracValid} & \textbf{FracUnique} & \textbf{FracPassesMCF} & \textbf{IntDiv} \\ \hline
\texttt{smiles-rnn-base}    &    $0.995$       &  $0.995$          &   $0.532$            &  $0.857$      \\ 
\texttt{smiles-rnn-mcf-dpo} &    $0.994$ (\textcolor{red}{$-0.1\%$})      &  $0.986$ (\textcolor{red}{$-0.9\%$})         &  $0.872$  (\textcolor{darkgreen}{$+65\%$})           &  $0.849$ (\textcolor{red}{$-0.9\%$})      \\ \hline\hline
\texttt{smiles-gpt-base}    &    $0.995$       &  $0.995$          &  $0.517$             &  $0.856$      \\
\texttt{smiles-gpt-mcf-dpo} &    $0.994$ (\textcolor{red}{$-0.1\%$})       &  $0.992$  (\textcolor{red}{$-0.3\%$})        &   $0.927$   (\textcolor{darkgreen}{$+79\%$})         &  $0.847$ (\textcolor{red}{$-1.1\%$})      \\ \hline
\end{tabular}}
\end{center}
\caption{\textbf{Results from MCF fine-tuning experiments.} We observe that for both architectures, DPO fine-tuning significantly increases the rate of filter passing while minimally affecting other metrics.}
\label{tab:mcf-results}
\end{table}

For our first set of experiments, we consider feedback in the form of a set of filters that exemplify the desiderata a medicinal chemist may want from generated molecules. Specifically for this task, molecules are assigned a binary score in $\{\texttt{PASS}, \texttt{FAIL}\}$ based on the following criteria. A molecule fails the filters if it does not satisfy any one of the following:
\begin{enumerate}[noitemsep, topsep=1pt]
    \item The molecule does not contain any one of a set of 91 distinct SMARTS filters, representing undesired chemical substructures (e.g., reactive substructures, difficult to synthesize, etc). 
    \item The molecule has a molecular weight between 300 and 600 Daltons.
    \item The molecule has fewer than 2 chiral centers.
    \item The molecule contains fewer than 8 rings.
\end{enumerate}
Molecules with the score \texttt{PASS} are considered positive examples for the DPO step, and molecules with the score \texttt{FAIL} are considered negative examples. Using the pre-trained \texttt{smiles-gpt} and \texttt{smiles-rnn} models, we sample 100,000 molecules, and perform this filtering in order to obtain training data for DPO. A schematic representation of this workflow is presented in Figure \ref{fig:schematic}.

We report results from these experiments in Table \ref{tab:mcf-results}, where \textbf{FracPassesMCF} denotes the fraction of the 10,000 generated molecules that pass the above-defined filters. We observe that the baseline models, \texttt{smiles-gpt-base} and \texttt{smiles-rnn-base}, attain filter passing rates of $52\%$ and $53\%$, respectively. After fine-tuning with DPO, these rates improve significantly---by $79\%$ and $65\%$, respectively. Moreover, these improvements come with little-to-no degradation in the other metrics.

\subsection{Predicted bioactivity}
\label{sec:EGFR-opt}
\begin{table}[]
\begin{center}
\resizebox{\textwidth}{!}{\begin{tabular}{|l|l|l|l|l|}
\hline
\textit{Model}                       & \textbf{FracValid} & \textbf{FracUnique} & \textbf{FracPredActive} & \textbf{IntDiv} \\ \hline
\texttt{smiles-rnn-chembl}    &    $0.955$       &  $0.942$          &   $0.096$            &  $0.865$      \\
\texttt{smiles-rnn-EGFR-dpo} &    $0.951$ (\textcolor{red}{$-0.4\%$})       &  $0.871$ (\textcolor{red}{$-7.5\%$})         &  $0.602$  (\textcolor{darkgreen}{$+527\%$})           &  $0.829$ (\textcolor{red}{$-4.2\%$})      \\ \hline\hline
\texttt{smiles-gpt-chembl}    &    $0.939$       &  $0.927$          &  $0.090$             &  $0.864$      \\
\texttt{smiles-gpt-EGFR-dpo} &    $0.871$ (\textcolor{red}{$-7.2\%$})      &  $0.763$ (\textcolor{red}{$-17.7\%$})         &   $0.790$   (\textcolor{darkgreen}{$+778\%$})         &  $0.799$ (\textcolor{red}{$-7.5\%$})      \\ \hline
\end{tabular}}
\end{center}
\caption{\textbf{Results from EGFR activity fine-tuning experiments.} We observe that for both architectures, DPO fine-tuning significantly increases the fraction of generated molecules predicted active against the protein target. In this case, however, we observe some degradation in the other metrics.}
\label{tab:egfr-results}
\end{table}
\begin{figure}[t!]
    \centering
    \includegraphics[scale=0.5]{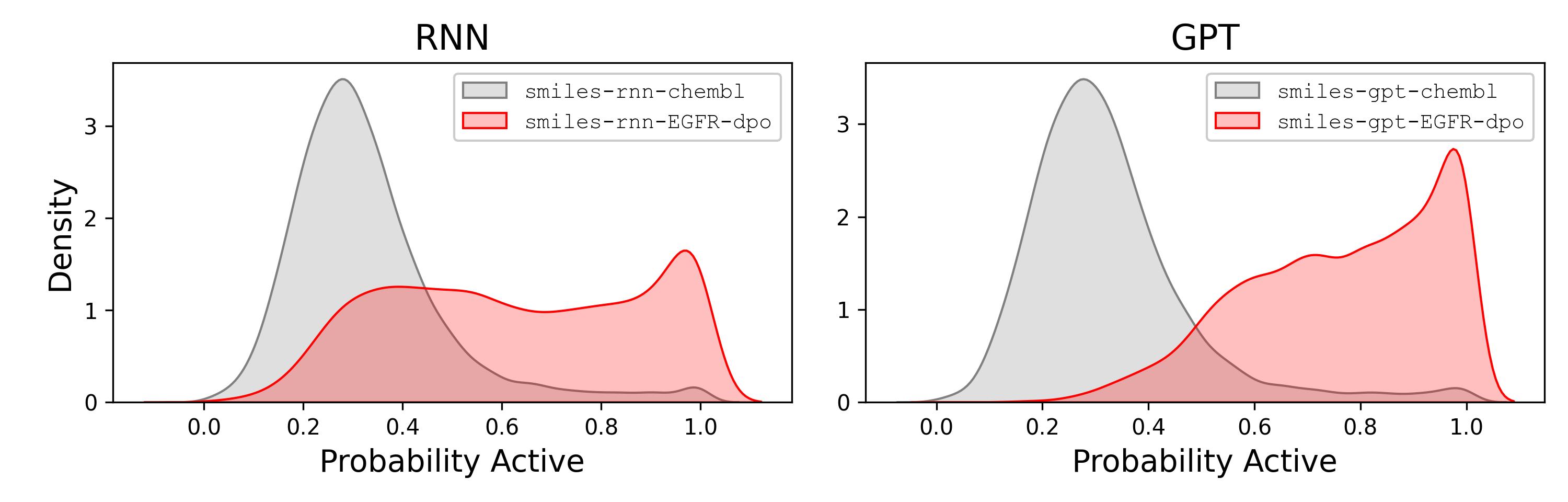}
    \caption{\textbf{Predicted probability of activity against EGFR for generated molecules.} After fine-tuning with DPO, the predicted probability of activity against EGFR for generated molecules is shifted significantly towards activity versus the baseline model.}
    \label{fig:EGFR-pred-prob}
\end{figure}

In our next set of experiments, we desire to sample molecules with high predicted bioactivity against a given target. Specifically, we focus on the protein kinase EGFR, which is known to be of clinical relevance for a number of diseases, primarily cancer \cite{EGFR}. Because the MOSES benchmark set doesn't contain kinase-inhibitor specific drugs, we first fine-tune the \texttt{smiles-rnn-base} and \texttt{smiles-gpt-base} models on a set of 115,000 kinase inhibitors extracted from the ChEMBL database, to obtain models which we call \texttt{smiles-rnn-chembl} and \texttt{smiles-gpt-chembl}, respectively. We then train a random forest binary classifier based on 3485 IC$_{50}$ measurements also extracted from ChEMBL. Here we consider a molecule active against the target if its IC$_{50}$ is less than 100nM, and inactive if its IC$_{50}$ is greater than 500nM\footnote{For the sake of the classifier considered here, we exclude compounds with ambiguous activity status.}. This results in 2568 active compounds, and 917 inactive. After holding out $15\%$ of examples for testing, the classifier achieves $95\%$ accuracy. Additional details can be found in Appendix \ref{sec:additional-egfr}.

Our goal in this section is to sample molecules that are predicted to be active against the EGFR target. We quantify this with the metric \textbf{FracPredActive}, measuring the fraction of sampled molecules that are predicted as active against EGFR by the binary classifier. In Table \ref{tab:egfr-results}, we observe that only approximately $9\%$ of the sampled molecules from \texttt{smiles-rnn-chembl} and \texttt{smiles-gpt-chembl} are predicted to be active. We then perform DPO fine-tuning by sampling 100,000 SMILES strings from each model, and labeling them as \texttt{ACTIVE} or \texttt{INACTIVE} using the trained classifier. These examples are then used as positive/negative examples for DPO, resulting in further fine-tuned models \texttt{smiles-rnn-EGFR-dpo} and \texttt{smiles-gpt-EGFR-dpo}. After fine-tuning with DPO, we find significant improvements in predicted activity, increasing by $527\%$ and $778\%$ for the RNN and GPT models, respectively. However, unlike in the filtering experiments, we observe the improvement in predicted activity comes at the expense of some degradation across the other metrics. We hypothesize that this is due to the limited number of positive examples for this task compared to the filtering experiments (only $\approx 9\%$ of the fine-tuning data versus $\approx 50\%$ for the filtering task).

To visualize the shift in the distribution of predicted activity, in Figure \ref{fig:EGFR-pred-prob} we plot the distributions of predicted probability of activity before and after fine-tuning. We observe that DPO effectively shifts the distribution towards higher activity for both architectures, using only synthetically generated data. 
\section{Conclusions}
\label{sec:conclusion}

We have studied the use of DPO to tune molecular language models in concordance with chemist preferences. Across both experimental settings evaluated here, we find that fine-tuning with DPO is very effective at directing generation towards molecules with desired properties. Moreover, we find that training with DPO is straightforward and computationally low-cost. 

Our preliminary investigations suggests many directions ripe for further study; we list a few of particular interest here.

\begin{itemize}[noitemsep, topsep=1pt]
    \item We observe in some (but not all) experiments that fine-tuning with DPO leads to lower diversity in generated molecules. We hypothesize that this may occur particularly when the labeled datasets do not contain sufficiently many positive examples. Are there ways to mitigate this issue?
    \item DPO requires that scores be effectively binarized into an ordering $y\succ y'$ (as we have done with bioactivity in Section \ref{sec:EGFR-opt}). However, many scores relevant for molecular generation tasks are more naturally represented as a continuous scalar, and binarizing these scores may result in loss of important information. Can the DPO technique be generalized to allow for continuous labels?
    \item As a preliminary investigation, we focused on preference labels that can be computed automatically. However, the recent success of fine-tuning from human feedback in language models suggests manually-curated labels from chemists may yield promising results.
\end{itemize}

\bibliographystyle{alpha}
\bibliography{references.bib}

\newpage
\appendix
\section{Additional experimental details}
\label{sec:additional-experimental-details}

\subsection{Training details.}
\label{sec:training-details}
Here we report details of our training procedure, both for pre-training and for fine-tuning. 

\paragraph{Architecture details.} The RNN model consists of 3 LSTM layers using an embedding dimension of 768, followed by a linear layer mapping to the vocabulary size. The implementation is based on the one provided in the MOSES benchmark suite \cite{MOSES}. The GPT model consists of 8 transformer blocks, each with 8 attention heads and an embedding dimension of 256, followed by a final classification head of the dimension of the vocabulary size. The implementation is closely based on minGPT \cite{minGPT}. 

\paragraph{Pre-training.} For pre-training the RNN model on MOSES, we use the Adam optimizer with an initial learning rate of 1e-3, decayed by a factor of $0.1$ every 10 epochs, for a total of 80 epochs. For the GPT model, we using the AdamW optimizer with an initial learning rate of 6e-4, cycled using a cosine annealing schedule for 40 epochs. The final models are selected based on the validation loss on the MOSES test set. 

\paragraph{Unsupervised fine-tuning.} During the unsupervised fine-tuning step (used for the \texttt{smiles-rnn-chembl} and \texttt{smiles-gpt-chembl} models), we use the same optimizers and number of epochs as in the pre-training stage, except for the GPT model we disable learning rate decay. The final models are selected based on the validation loss on a hold-out set of 15,000 SMILES strings. 

\paragraph{DPO.} Our implementation of DPO closely mirrors the original implementation provided by the authors in \cite{DPO}. We use the RMSProp optimizer with a Lambda learning rate scheduler (as implemented in PyTorch \cite{pytorch}) for 80 epochs. 

\subsection{Molecular filtering}
\label{sec:additional-filtering}
\paragraph{SMARTS filters.} The SMARTS filters we use are included for reference in Table \ref{tab:SMARTS}, along with a column "count" indicating the minimum number of times the pattern can be observed for the molecule to fail the filters. 

\subsection{EGFR activity optimization}
\label{sec:additional-egfr}
\paragraph{Binary classifier.} The binary classifier is trained using the default parameters of the \texttt{RandomForestClassifier} class in \texttt{scikit-learn} \cite{scikit-learn}. To featurize molecules, we use $1024$-bit Morgan fingerprints, which we compute using the \texttt{RDKit} package in Python. As we mention in Section \ref{sec:EGFR-opt}, we hold out $15\%$ of our data for testing, resulting in $523$ testing examples. On this hold-out set, the model achieves a raw accuracy of $95\%$; we report additional performance metrics in Table \ref{tab:binary-classifier-performance}. We remark that the classifier used here is largely for demonstration, and that, depending on the scenario of interest, better predictors may be available. 

\begin{table}[]
\begin{center}
\begin{tabular}{|l|l|l|l|l|}
\cline{1-5}
                               & \textit{precision} & \textit{recall} & \textit{f1-score} & \textit{support} \\ \hline
\textbf{inactive} & 0.89      & 0.92   & 0.90     & 126     \\ 
\textbf{active}   & 0.97      & 0.96   & 0.97     & 397     \\ \hline
\end{tabular}
\end{center}
\caption{Test set performance of EGFR binary classifier.}
\label{tab:binary-classifier-performance}
\end{table}

\subsection{Scaffold conditioning}
\label{sec:additional-scaffold-conditioning}

\section{Model samples}

Here we include some samples from the molecular language models referenced throughout the paper. 

\begin{figure}
    \centering
    \includegraphics[scale=0.5]{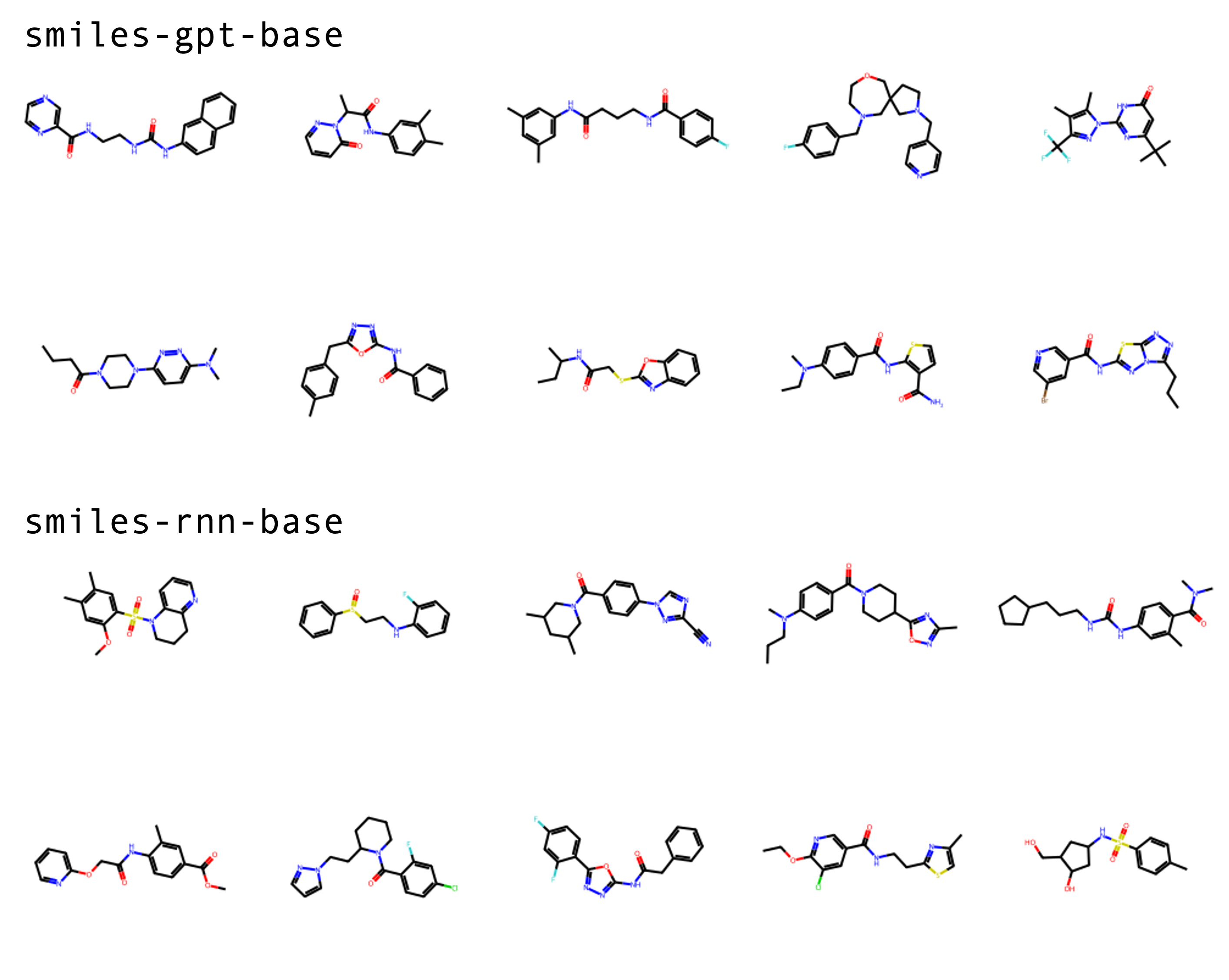}
    \caption{Samples from baseline GPT and RNN models, pre-trained on MOSES.}
    \label{fig:smiles-base-samples}
\end{figure}

\begin{figure}
    \centering
    \includegraphics[scale=0.5]{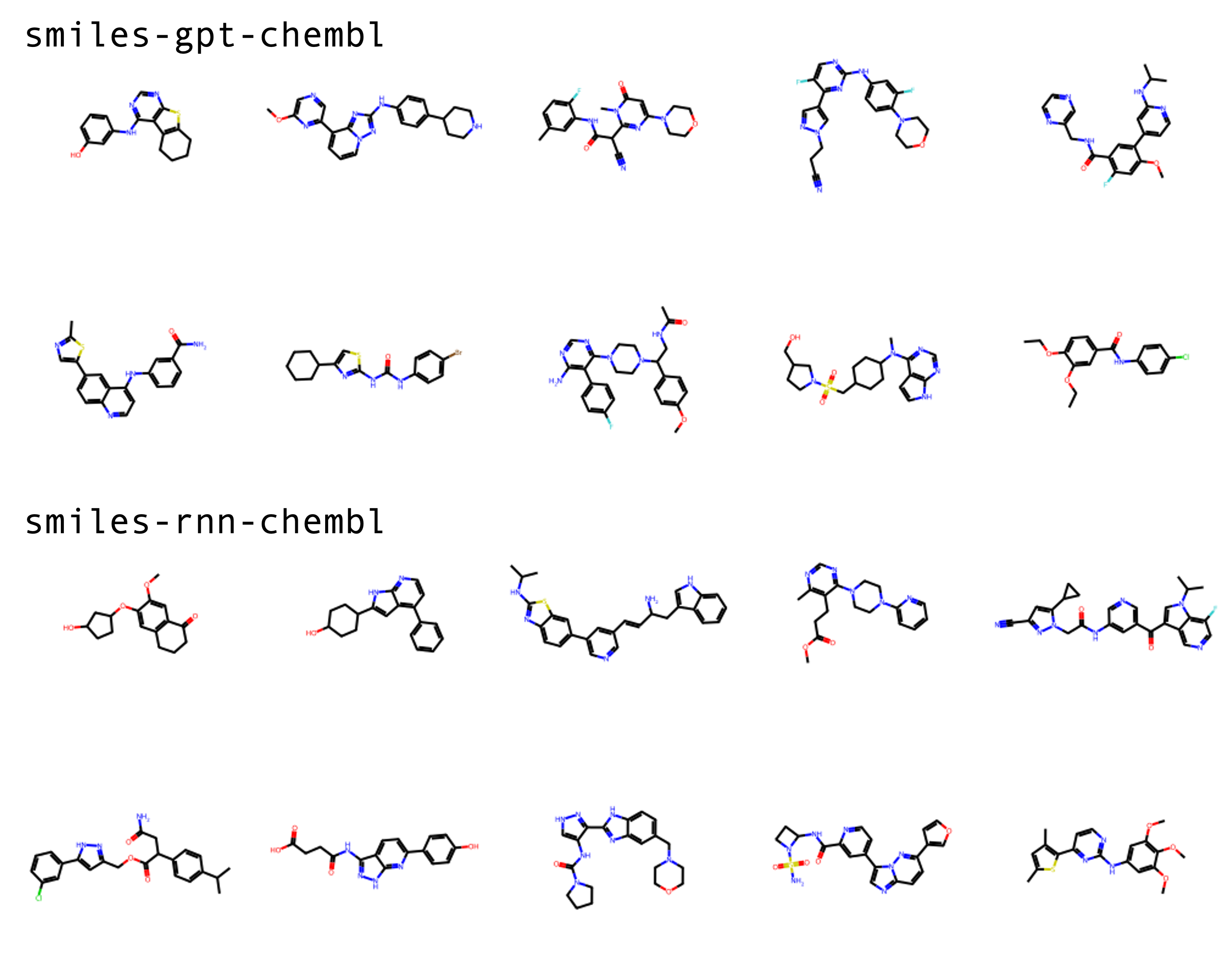}
    \caption{Samples from ChEMBL GPT and RNN models, pre-trained on MOSES, and fine-tuned on ChEMBL set of kinase inhibitors.}
    \label{fig:smiles-chembl-samples}
\end{figure}

\begin{figure}
    \centering
    \includegraphics[scale=0.5]{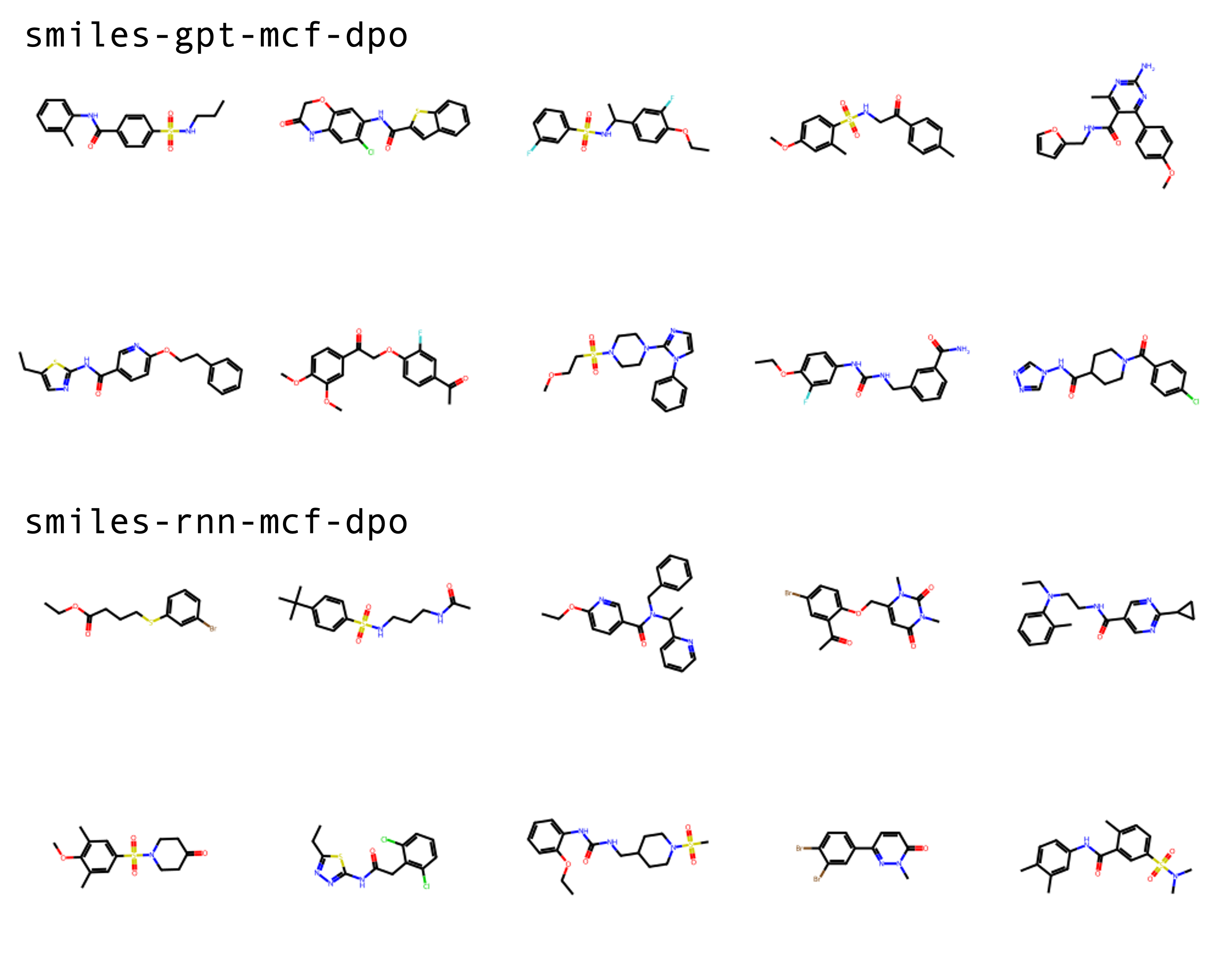}
    \caption{Samples from ChEMBL GPT and RNN models, pre-trained on MOSES, and fine-tuned using DPO using medicinal chemist filtering, as described in Section \ref{sec:MCF-opt}.}
    \label{fig:smiles-mcf-dpo-samples}
\end{figure}

\begin{figure}
    \centering
    \includegraphics[scale=0.5]{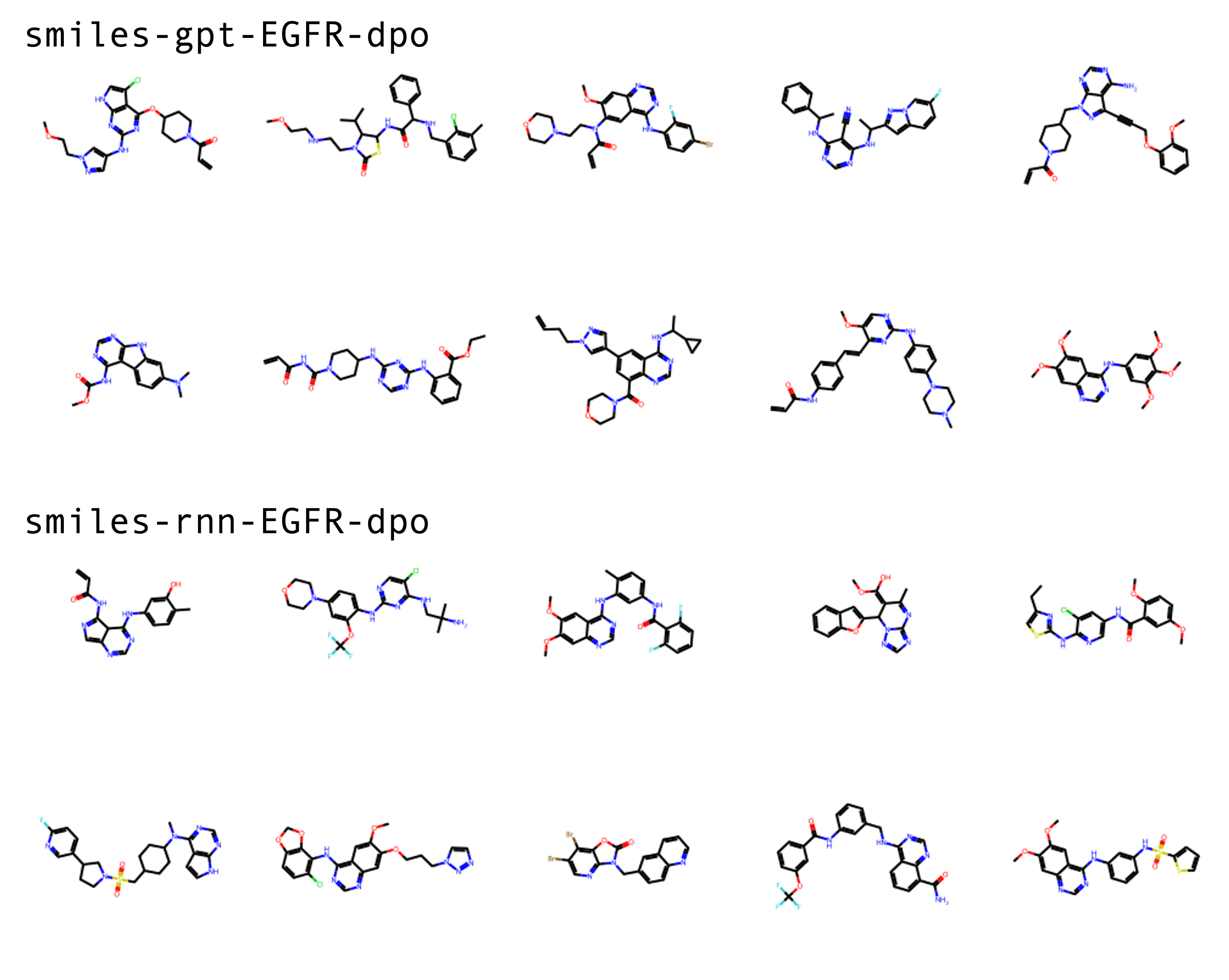}
    \caption{Samples from ChEMBL GPT and RNN models, pre-trained on MOSES, fine-tuned on ChEMBL, and further fine-tuned using DPO using predicted activity against EGFR, as described in Section \ref{sec:EGFR-opt}.}
    \label{fig:smiles-EGFR-dpo-samples}
\end{figure}

\newpage 

\begin{longtable}{| p{.85\textwidth} | p{.15\textwidth} |} 
\hline
\textbf{SMARTS} & \textbf{Count} \\
\hline
\texttt{*-N=[N+]=[N-]} & 1 \\ \texttt{BrccBr} & 1 \\ \texttt{C(-C\#N)(-C\#N)} & 1 \\ \texttt{C(-C\#N)C(-C\#N)} & 1 \\ \texttt{C1(=O)[C,c]:[C,c]C(=O)[C,c][C,c]1} & 1 \\ \texttt{C14~*~*~*~*~C~1~*~*~C2~C3~*~*~*~C~3~*~*~C~2~4} & 1 \\ \texttt{C~1~C~C~2~C~C~[\#6]~[\#6]~C~2~C~C~1} & 1 \\ \texttt{N=[N;!R]} & 1 \\ \texttt{NC\#[NX1]} & 1 \\ \texttt{S\#N} & 1 \\ \texttt{S(=[N])} & 1 \\ \texttt{S[N;!R][N,n]} & 1 \\ \texttt{[!\$(C=O)][NX3]-[OH]} & 1 \\ \texttt{[!\$(C=O)][n,N]-[OH]} & 1 \\ \texttt{[\#6;!R][CX3;\$([!R][\#6])](=[O])[\#6;!R][\#6](=[O])} & 1 \\ \texttt{[\#6]1=[\#6]-[\#6]-1} & 1 \\ \texttt{[\#6]1=[\#6]-[\#7,\#6]=[\#6]-1} & 1 \\ \texttt{[\#6][\#16][\#16][\#6]} & 1 \\ \texttt{[\#7,S][C;!R](=S)N} & 1 \\ \texttt{[\#7]~[\#6]([Cl,Br])(~[\#7])} & 1 \\ \texttt{[\#8,\#7,\#16]-[CH,CH2;!R]-[\#8,\#7,\#16][CX4,c;!\$(C(=O))]} & 1 \\ \texttt{[\#8;!\$(O[CH2][CH2][O,N])]} & 6 \\ \texttt{[\$([\#7+][OX1-]),\$([\#7v5]=[OX1]);!\$([\#7](~[O])~[O]);!\$([\#7]=[\#7])]} & 1 \\ \texttt{[\$([CX3]=[CX3]([H,!C])[CX2]\#[NX1]);!R]} & 1 \\ \texttt{[\$([HO]-N-C(=O))]} & 1 \\ \texttt{[\$([N]=[CX3]c);!R]} & 1 \\ \texttt{[*r5R1]1[CR2]2[CR1][CR1][CR1][CR1][CR2]2[*r5R1][*r5R1]1} & 1 \\ \texttt{[B;!\$([B][OX1])]} & 1 \\ \texttt{[B]([OX2]C)([OX2][C])} & 1 \\ \texttt{[C,O][N+,N](=[OX1])} & 1 \\ \texttt{[C;H1](=O)} & 1 \\ \texttt{[CH2]=[CH]-[N,O,S]} & 1 \\ \texttt{[CH]([OH])[CH]([OH])[CH]([CH2][OH])} & 1 \\ \texttt{[CH]([OH])[CH]([OH])[CH]([OH])} & 1 \\ \texttt{[CH]=[CH]-N(=O)(~O)} & 1 \\ \texttt{[C]!@[C]!@[C]!@[C]!@[C]} & 1 \\ \texttt{[C](=[O,S])[CH2][Br,Cl,I]} & 1 \\ \texttt{[C]([Br,Cl])([Br,Cl])} & 1 \\ \texttt{[C]=[\#16+][O-]} & 1 \\ \texttt{[C]=[CH]-[O,NH,NH2]} & 1 \\ \texttt{[Cl,Br,I]} & 5 \\ \texttt{[I,Se,se,Si,P]} & 1 \\ \texttt{[I,Se,se,Si]} & 1 \\ \texttt{[N!\$(N-O)]=O} & 1 \\ \texttt{[N+,N](=[OX1])[OX1]} & 2 \\ \texttt{[N+,n+;!\$([N+,N](=[OX1])[OX1])][C,c]} & 1 \\ \texttt{[N+]\#[C-]} & 1 \\ \texttt{[N+](=O)[O-]} & 1 \\ \texttt{[N,!R][C](=[N,S])} & 1 \\ \texttt{[N,O,S][F,Cl,Br]} & 1 \\ \texttt{[N,O]([SX2,SX3])} & 1 \\ \texttt{[N,S,O,Br][C]=[\#6]} & 1 \\ \texttt{[N,S][c,C]([SH,S-1])[\#7]} & 1 \\ \texttt{[N,n](-[O])} & 1 \\ \texttt{[N,n]([O])} & 1 \\ \texttt{[N;!\$(N=O)]-O} & 1 \\ \texttt{[N;!\$(N=O)]-[OH]} & 1 \\ \texttt{[NX3;!R][NX2]=[*]} & 1 \\ \texttt{[NX3][CH2][CH2][Cl,Br]} & 1 \\ \texttt{[NX3][NX3;!R]} & 1 \\ \texttt{[N]-[NN;R]} & 1 \\ \texttt{[N]-[n]} & 1 \\ \texttt{[O,N,S,Cl,F,Br][CH,CH2][O,N,S,Cl,Br][CX4,c;!\$(C(=O))]} & 1 \\ \texttt{[O,N,S][CH,CH2;!R][O,N,S][CX4,c;!\$(C(=O))]} & 1 \\ \texttt{[O][O]} & 1 \\ \texttt{[PH]([c,C,N])} & 1 \\ \texttt{[PX4D4]([OX1,N])([O,N])} & 1 \\ \texttt{[P](=O)([O,N])} & 1 \\ \texttt{[P]([O,N,C,S])} & 1 \\ \texttt{[S+1;!\#16\#8]} & 1 \\ \texttt{[S,C;+1]} & 1 \\ \texttt{[S,C](=[O,S])[F,Br,Cl,I]} & 1 \\ \texttt{[S;!\$(S[OX1-1])]} & 3 \\ \texttt{[S;D4](=[OX1])(=[OX1])[O;H1,O-]} & 1 \\ \texttt{[SH]} & 1 \\ \texttt{[S][S]} & 1 \\ \texttt{[Se][Se]} & 1 \\ \texttt{[c]1C(=O)[NX3][CH2][c]1} & 1 \\ \texttt{[c]1C(=O)[NX3][C](=O)[c]1} & 1 \\ \texttt{[c]1[c][s,n,o,c][s,n,o,c]1} & 1 \\ \texttt{[n+]-[O-]} & 1 \\ \texttt{[o,O;+1]} & 1 \\ \texttt{[r3O,r3N,r3S]} & 1 \\ \texttt{[r7]} & 1 \\ \texttt{[r{8-}]} & 1 \\ \texttt{a-[C]=[CH2]} & 1 \\ \texttt{a[CH,CH2][Cl,Br]} & 1 \\ \texttt{a[CH2][CH2;!R][CH2][CH2]C*} & 1 \\ \texttt{a[C]=[CH2]} & 1 \\ \texttt{a[OH]} & 3 \\ \texttt{c[Br]n} & 1 \\ 
\hline
\caption{SMARTS patterns used for filtering in Section \ref{sec:MCF-opt}.} 
\label{tab:SMARTS}
\end{longtable}

\end{document}